\pdfoutput=1

\documentclass[11pt]{article}

\usepackage{acl}

\usepackage{times}
\usepackage{latexsym}

\usepackage[T1]{fontenc}

\usepackage[utf8]{inputenc}

\usepackage{microtype}

\usepackage{amsmath}
\usepackage{amsfonts}
\usepackage{booktabs}
\usepackage{multicol}
\usepackage[weather]{ifsym}
\usepackage{bbding} 

\usepackage{enumitem}

\usepackage{xspace}
\xspaceaddexceptions{$}

\usepackage{adjustbox}
\usepackage[normalem]{ulem}
\usepackage{pifont}
\usepackage{stfloats}
\usepackage{cancel}

\usepackage{color, colortbl}
\definecolor{LightCyan}{rgb}{0.88,1,1}

\newcommand\xoutpars[1]{\let\helpcmd\xout\parhelp#1\par\relax\relax}
\newcommand\soutpars[1]{\let\helpcmd\sout\parhelp#1\par\relax\relax}
\long\def\parhelp#1\par#2\relax{%
  \helpcmd{#1}\ifx\relax#2\else\par\parhelp#2\relax\fi%
}

\definecolor{cadmiumgreen}{rgb}{0.0, 0.42, 0.24}
\definecolor{goldenpoppy}{rgb}{0.99, 0.76, 0.0}
\definecolor{goldenbrown}{rgb}{0.6, 0.4, 0.08}
\definecolor{goldenrod}{rgb}{0.85, 0.65, 0.13}

\newcommand{\xstar}{\textcolor{goldenpoppy}{\ding{72}}}

\newcommand{\smatch}[0]{\textsc{Smatch}\xspace}
\newcommand{\fgsmatch}[0]{\textsc{FineSmatch}\xspace}
\newcommand{\wlk}[0]{\textsc{WlKernel}\xspace}
\newcommand{\wwlk}[0]{\textsc{WwlKernel}\xspace}

\newcommand{\sbert}[0]{SBERT\xspace}

\newcommand{\sbertx}[0]{S$^3$BERT\xspace}
\newcommand{\zielscheibe}[0]{\includegraphics[scale=0.5]{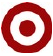}}
\newcommand{\zahnrad}[0]{\includegraphics[scale=0.5]{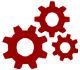}}

\title{SBERT studies Meaning Representations: Decomposing Sentence Embeddings into Explainable Semantic Features}

\author{Juri Opitz \\
  Dept.\ of Computational Linguistics \\
  Heidelberg University \\
  69120 Heidelberg \\
  \texttt{opitz.sci@gmail.com} \\\And
  Anette Frank \\
  Dept.\ of Computational Linguistics \\
  Heidelberg University \\
  69120 Heidelberg \\
  \texttt{frank@cl.uni-heidelberg.de} \\}

\begin{document}
\maketitle
\begin{abstract}
Models based on large-pretrained language models, such as S(entence)BERT, provide effective and efficient sentence embeddings that show high correlation to human similarity ratings, but lack interpretability. On the other hand, graph metrics for graph-based meaning representations (e.g., Abstract Meaning Representation, AMR) can make explicit the semantic aspects in which two sentences are similar. However, such metrics tend to be slow, rely on parsers, and do not reach state-of-the-art performance when rating sentence similarity. 

In this work, we aim at the best of both worlds, by learning to induce \underline{S}emantically \underline{S}tructured \underline{S}entence BERT embeddings (\sbertx). Our \sbertx embeddings are composed of explainable sub-embeddings that emphasize various semantic sentence features (e.g., semantic roles, negation, or quantification). We show how to i) learn a decomposition of the sentence embeddings  into semantic features, through approximation of a suite of interpretable AMR graph metrics, and how to ii) preserve the overall power of the neural embeddings by controlling the decomposition learning process with a second objective that enforces consistency with the similarity ratings of an SBERT teacher model. In our experimental studies, we show that our approach offers interpretability -- while fully preserving the effectiveness and efficiency of the neural sentence embeddings.
\end{abstract}

\section{Introduction}

Abstract Meaning Representation (AMR) represents the meaning of a sentence as a directed, rooted and acyclic graph \cite{banarescu-etal-2013-abstract}. It shows events and entities referred to in a sentence, their semantic roles and key semantic relations such as \textit{cause, time, purpose, instrument, negation}. 

The explicit representation of meaning in AMR has motivated research into AMR metrics that measure meaning similarity of the underlying sentences. E.g., AMR metrics are used for semantics-focused NLG evaluation \cite{opitz2020towards, manning-schneider-2021-referenceless, zeidler-etal-2022-dynamic}, a semantic search engine \cite{bonial-etal-2020-infoforager}, comparison of cross-lingual AMR \cite{uhrig-etal-2021-translate, wein-etal-2022-effect}, and argument similarity \cite{opitz-etal-2021-explainable}. Moreover, fine-grained AMR metrics can assess meaning similarity of semantic sub-aspects that AMR explicitly captures, e.g., semantic roles or negation \cite{damonte-etal-2017-incremental}.

However, when measuring similarity rating performance against human ratings in the typical zero-shot setting on tasks like STS \cite{baudivs2016sentence} or SICK \cite{marelli-etal-2014-sick}, the (untrained) AMR metrics tend to lag behind large models such as \sbert \cite{reimers-gurevych-2019-sentence} that computes sentence embeddings with a Siamese BERT transformer model \cite{devlin-etal-2019-bert}. 

Notably, \sbert alleviates the need for end-to-end similarity inference on each sentence pair. Instead, it infers the embedding of each sentence individually, and calculates similarity with simple vector algebra, which greatly reduces clustering and search time. AMR metrics, by contrast, tend to be slower, are often NP-hard \cite{cai-knight-2013-smatch} and rely on a parser.

Hence, we find complementarity in these two approaches of rating sentence similarity: AMR metrics offer high explainability -- but tend to be slow and need improvement to compete in benchmarking. By contrast, neural embeddings show strong empirical performance and efficiency -- but lack explainability. 

Aiming at the best of these worlds, we propose to leverage multi-aspect AMR metrics as a means to teach a pre-trained \sbert model on how to structure its sentence embedding space such that it explicitly captures specific abstract aspects of meaning similarity, in terms of semantic roles, negation, quantification, etc. This has to be undertaken with care, to prevent catastrophic forgetting \cite{goodfellow2013empirical, 10.1007/978-3-030-58598-3_28}, which could negatively impact \sbert's empirical performance and the overall effectiveness of its embeddings. 

Our contributions:

\begin{enumerate}[label=\arabic*., noitemsep]
    \item To increase the explainability of sentence embeddings, we propose a method that performs \textit{Semantic Decomposition} in the SBERT sentence embedding space, to yield S$^3$BERT (\underline{S}emantically \underline{S}tructured \underline{S}BERT) embeddings. 
    \sbertx sub-embeddings express key semantic sentence features that reflect AMR metric measurements taken on the sentences' underlying meaning representations.
    \item To prevent catastrophic forgetting, we include a consistency objective that controls the decomposition learning process and projects important semantic information not captured by AMR to a residual sub-embedding. 
    \item Our experiments and analyses in zero-shot sentence and argument similarity tasks show that \sbertx embeddings are more explainable than \sbert embeddings while fully preserving \sbert's efficiency and accuracy. 
    \item Code and data are publicly released: \url{https://github.com/flipz357/S3BERT}
\end{enumerate}

\section{Related work}

\paragraph{\sbert and friends: High efficacy at the cost of lower interpretability} Since its introduction by \citet{reimers-gurevych-2019-sentence}, S(entence)BERT has become a popular method for computing sentence similarity \cite{thakur2020augmented, reimers-gurevych-2020-making, sbert-wk, sbert-TA}. This is due to two key properties: \sbert shows strong results on similarity benchmark tasks and it is highly efficient. E.g., it allows rapid sentence clustering since the BERT backbone is called independently for each sentence, alleviating the need for pair-wise model inferences.

However, \sbert provides little explainability. While different linguistic indicators have been identified for or within BERT \cite{jawahar-etal-2019-bert-structure, lepori-mccoy-2020-picking-structure, warstadt-etal-2019-investigating-structure, puccetti-etal-2021-bert-stucture}, this insight by itself does not provide us with any rationale for high (or low) sentence similarity in specific cases, and so, to achieve \textit{local} explainability \cite{danilevsky-etal-2020-survey}, we would have to, at least, analyze attention weights \cite{clark-etal-2019-bert, wiegreffe-pinter-2019-attention} or gradients \cite{selvaraju2017grad, sanyal-ren-2021-discretized, bastings-filippova-2020-elephant} of regions associated with linguistic properties. But even then, it can be unclear how exactly to interpret the results  \cite{jain-wallace-2019-attention, wiegreffe-pinter-2019-attention, wang-etal-2020-gradient, ferrando-costa-jussa-2021-attention-weights}. In a different direction, \citet{kaster-etal-2021-global} aim to explain BERTscore \cite{Zhang*2020BERTScore} predictions with a regressor. But unlike other explanation methods, this approach is detached from the underlying BERT model and may suffer from indirection effects. Instead, we target local self-\-ex\-plain\-ability \cite{danilevsky-etal-2020-survey} by structuring \sbert's sentence embedding space into subspaces that emphasize explicit facets of meaning. Parts of this idea are inspired from \citet{rothe-schutze-2016-word}, who compose four semantic spaces of \textit{word vectors}, using a lexical resource. Without such a resource, and targeting sentence embeddings, we aim to leverage and structure semantic knowledge already present in the model, while injecting new knowledge that we obtain from metrics grounded in a multi-faceted theory of meaning, namely AMR.

\paragraph{AMR metrics: the cost of interpretability} AMR graphs \cite{banarescu-etal-2013-abstract} explicate aspects of meaning, such as entities, events, coreference, or negation. Metrics defined over AMRs therefore show specific aspects in which two sentences are similar or different, which makes them attractive for tasks going beyond parser evaluation, such as NLG evaluation \cite{opitz2020towards, manning-schneider-2021-referenceless}, semantic search \cite{bonial-etal-2020-infoforager}, explainable argument similarity rating \cite{opitz-etal-2021-explainable}, or investigation of cross-lingual divergences \cite{uhrig-etal-2021-translate, wein-etal-2022-effect}. While classical AMR metrics assess semantic similarity structurally via binary matches of triples \cite{cai-knight-2013-smatch}, recent metrics target larger contexts and graded similarity scoring \cite{opitz-tacl, opitz2021weisfeiler}, e.g., to match a subgraph \textit{cat :mod young} against a node \textit{kitten}.

But this high degree of explainability comes at a price: AMR metrics tend to be slow since they i) compute costly graph alignments \cite{cai-knight-2013-smatch} and/or ii) require AMR parsers \cite{opitz2022smaragd} that are typically slow due to auto-regressive inference of large LMs \cite{T5, lewis2019bart}. iii) They are untrained, and thus tend to lag behind \sbert-based metrics in empirical settings \cite{opitz2021weisfeiler}. We aim to overcome these weaknesses by making sentence embeddings capable of expressing AMR metrics while preserving the full power of neural sentence embeddings. 

\paragraph{Sentence and argument similarity} Several works and resources aim to capture human sentence similarity ratings. E.g., SICK \cite{marelli-etal-2014-sick} rates \textit{semantic relatedness} and STS \cite{baudivs2016sentence} \textit{semantic similarity}, on 5-point Likert scales. \textit{Relatedness} and \textit{Similarity} have been argued to be very similar notions, albeit not the exact same \cite{budanitsky2006evaluating, kolb-2009-experiments}.\footnote{Only the highest rating on the SICK and STS Likert scales mean the exact same: two sentences are equivalent in meaning.}

An emergent branch of sentence similarity is the similarity of natural language arguments \cite{reimers-etal-2019-classification, opitz-etal-2021-explainable, behrendt-harmeling-2021-arguebert}, which finds broad application scenarios, e.g., in argument search engines \cite{maturana1988reality, wachsmuth2017building, ajjour2019data, mirko2020towards, slonim2021autonomous}.

While much research has been devoted to improving the accuracy of similarity rating systems, little attention has been paid to uncovering the features that  (in the eyes of a human) make two sentences similar or dissimilar \cite{zeidler-etal-2022-dynamic}. In our work, we propose a method that can potentially help uncover such features, while provably preserving strong rating accuracy.

\section{From \sbert to \sbertx: Structuring embedding space with AMR}
\label{sec:model}

\paragraph{Preliminary I: \sbert sentence embeddings and similarity} Let $SB$ be a function that maps an input sentence $s$ to a vector $e \in \mathbb{R}^d$. Given two sentence vectors $e = SB(s)$ and $e' = SB(s')$, we can compute, e.g., the cosine similarity of sentences: 

\begin{equation}
\label{eq:cosim}
   sim(e,e') = \frac{e^Te'}{|e||e'|}.
\end{equation}

\paragraph{Preliminary II: AMR and AMR metrics} An AMR $a \in A$ represents the meaning of a sentence in a directed acyclic graph. The AMR graph makes key aspects of meaning explicit, e.g., semantic roles or negation. Hence, given a pair of AMR graphs $\langle a,a'\rangle \in A \times A$, an AMR metric can measure \textit{overall} graph similarity, or similarity with respect to \textit{specific aspects}. We denote such a metric as

\begin{equation}
\label{eq:amrmetric}
    m^k: A \times A \rightarrow [0, 1],
\end{equation}

where $k$ indicates a particular semantic aspect, in view of which the graphs' similarity  
is assessed, e.g. negation. The AMR metrics we will apply in our work will be described in more detail in \S \ref{sec:setup}.

\begin{figure*}
    \centering
    \includegraphics[width=0.95\linewidth]{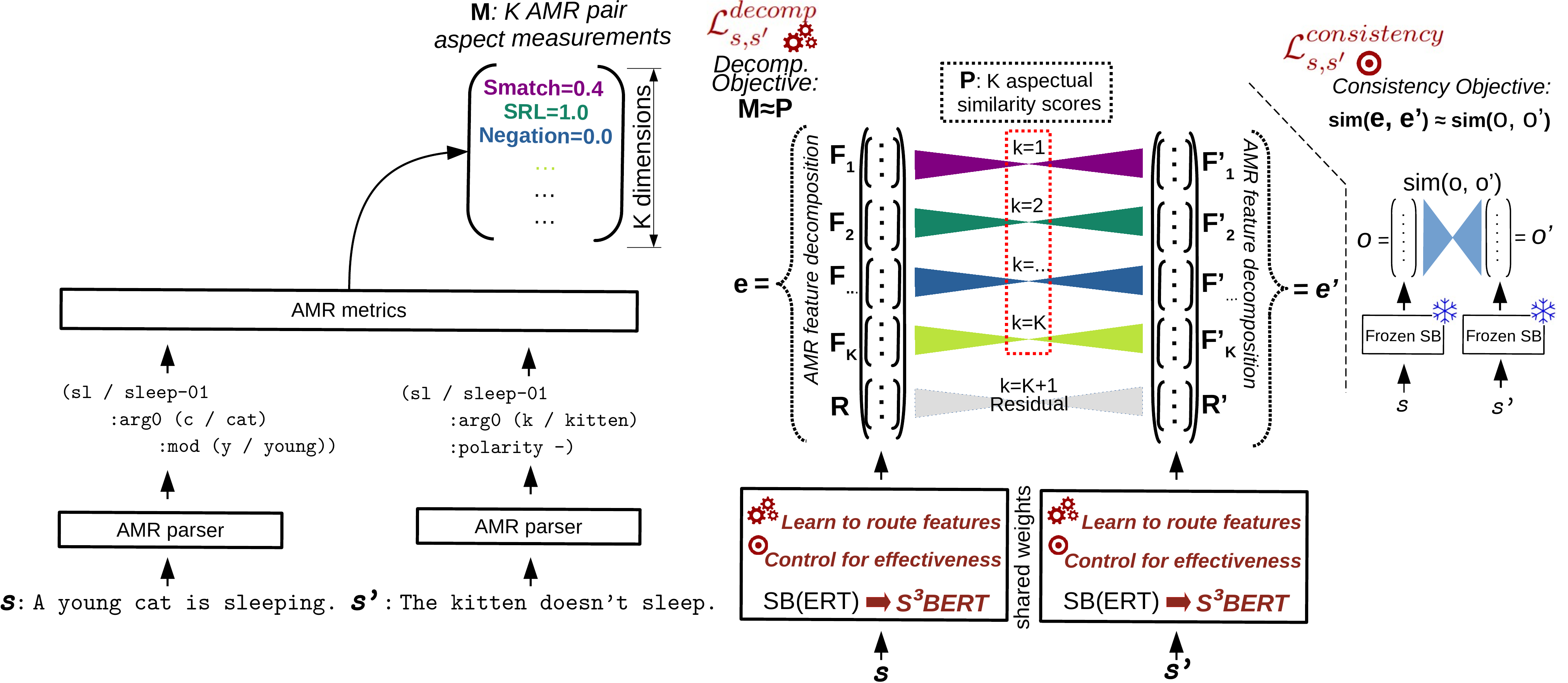}
    \caption{Overview of approach. \zahnrad~The decomposition objective structures the sentence embedding space into AMR sentence features (F$_1$...F$_K$): The process is guided by AMR metric approximation, through which \sbertx learns to disentangle and route the features. \zielscheibe~The consistency objective is aimed at preventing catastrophic forgetting: To preserve the overall effectiveness of the neural sentence embeddings, it controls the decomposition learning process and helps modeling the residual (R).
    }
    \label{fig:overview}
\end{figure*}

\subsection{Partitioning sentence embeddings into meaningful semantic AMR aspects} 

\paragraph{Problem statement} 
We aim to shape \sbert sentence embeddings in such a way that different sub-embeddings represent specific meaning aspects. This process of \textit{sentence embedding decomposition} is illustrated in Fig.\ \ref{fig:overview} (right): \sbert produces two embeddings $e$ and $e'$ that consist of sub-embeddings $F_1...F_K, R$ and $ F'_1...F'_K, R'$. E.g., $F_k$ may express negation features, while $F_z$ expresses semantic role features of a sentence. The residual $R$ offers space to model sentence features not covered by the pre-defined set of semantic features. 

Having established such decompositions, we can compute, e.g., sentence similarity with respect to semantic roles ($k=SRL$) by choosing subspaces $F_{SRL} \subset e=SB(s)$ and $F_{SRL}' \subset e'=SB(s')$, and calculating $sim(F_{SRL}, F_{SRL}')$ on the subspaces. This is indicated as \includegraphics[scale=0.4]{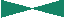} in Fig.\ \ref{fig:overview}. 

\paragraph{Assigning embedding dimensions to features} For convenience, let $i: \{1...K\} \rightarrow [0, d] \times [0, d]$ denote an AMR aspect-embedding assignment function where $d$ is the dimension of the (full) sentence embedding. This allows us to map any semantic category to a range of specific sentence embedding indices. E.g., a $h$-dimensional embedding for SRL sentence features for a sentence $s$ can be accessed via $SB(s)_{i(SRL)}$, where $v_{(start, end)}$ yields all dimensions from $start$ to $end$ of a vector $v$. Since we aim at a non-overlap decomposition, we ensure that $i(k) \cap i(k') \neq \emptyset \iff k = k'$.

\subsection{Learning to partition the semantic space} 

We presume that \sbert already contains some semantic features in some embedding dimensions. Hence, we want to achieve an arrangement of the embedding space according to our pre-defined partitioning, but also give it the chance to instill new knowledge about AMR semantics. 

In addition, to preserve \sbert's high accuracy, we aim to control the decomposition process in a way that lets us route internal semantic knowledge \textit{not} captured by AMR to the residual embedding. To this end, we propose a two-fold objective: \textit{Score decomposition} and \textit{Score consistency}.

\paragraph{Composing \sbertx score from AMR metrics} We build an AMR metric target $\mathbf{M}$ as shown in Fig.\ \ref{fig:overview} (left). Two AMRs, constructed from two sentences, are assessed  with AMR metrics in $K$ semantic aspects (Eq.\ \ref{eq:amrmetric}) yielding $\mathbf{M} \in \mathcal{M} = \mathbb{R}^K$. Additionally, let \textbf{P} be \sbertx's AMR metric predictions, i.e., $\mathbf{P} = [sim(F_1, F'_1), ... , sim(F_K, F'_K)]$.

For a training instance $(s, s', \mathbf{M})$, we calculate the following decomposition loss:

\vspace*{-5mm}
\begin{align}
\label{eq:disloss}
&\mathcal{L}^{decomp}_{s, s'} =\\ &\frac{1}{K}\sum_{k=1}^K \bigg[ \mathbf{M}_k - \beta^k \underbrace{sim(SB(s)_{i(k)}, SB(s')_{i(k)})}_{\mathbf{P}_k}\bigg]^2,     \nonumber
\end{align}

with $\beta^k$ a learnable scalar for easier projection onto a specific AMR metric's scale. The objective is also outlined as \textbf{P}$\approx$\textbf{M} in Fig.\ \ref{fig:overview}. 

Note that AMR graphs and metrics are only needed for training, not for inference.

\subsection{Preventing catastrophic forgetting} When training \sbertx only with the \textit{decomposition objective} (Eq.\ \ref{eq:disloss}), there is a great risk it will unlearn important information, since it is unrealistic to expect that sentence similarity can be \textit{fully} composed from the $K$ aspects measured by AMR metrics. It is also known that AMR metrics lag behind SBERT models in similarity rating accuracy. Hence, we control the decomposition learning process to include a $residual$ sub-embedding, to rescue important parts of semantic information not captured by AMR and AMR metrics. To this end, we propose a \textit{consistency objective}. 

Given a frozen \sbert ($SB^\text{\textcolor{blue}{\SnowflakeChevron}}$), and a training example $(s, s')$:

\begin{align*}\mathcal{L}^{consistency}_{s, s'} = & \bigg(sim(SB^\text{\textcolor{blue}{\SnowflakeChevron}}(s),  SB^\text{\textcolor{blue}{\SnowflakeChevron}}(s')) \\[-3mm] &- sim(SB(s), SB(s'))\bigg)^2.
\end{align*}

I.e., the control is established by imposing that \sbertx's overall similarity ratings be in accordance with a frozen \sbert's original ratings, but otherwise leaving freedom for the choice of structure in \sbertx's embedding space. Given independence of pairwise-targets, we can compute the loss efficiently on $b^2$ examples in batches of size $b$.

\subsection{Global objective} We finally combine the \textit{consistency objective} and the \textit{decomposition objective}. The cumulative loss for a batch $B=\{(S_i, S'_i, \mathcal{M}_i)\}_{i=1}^b$ is
\vspace*{-3mm}

\begin{equation}
\mathbf{L} = \frac{\alpha}{b} \sum_{i=1}^b \mathcal{L}^{decomp}_{S_i, S'_i} + \frac{1}{b^2} \sum_{i=1}^b \sum_{j=1}^b \mathcal{L}^{consistency}_{S_i, S'_j},
\end{equation}

where $\alpha$ weighs the two parts (we use $\alpha=1$).

\section{AMR metrics and data construction}
\label{sec:setup}

In Section \ref{sec:model}, Eq.\ \ref{eq:amrmetric}, we formally described an AMR metric. Now we consider the concrete metric instances we will use for \sbertx decomposition. We distinguish \textit{general} metrics that assess global AMR graph similarity, and \textit{aspectual} metrics that aim at assessing AMR similarity with respect to specific semantic categories, e.g., semantic roles.

\subsection{Global AMR similarity}
\label{sec:genamrsim}

\paragraph{\smatch} assesses the structural overlap of two semantic AMR graphs. It computes a best fitting combinatorial alignment between AMR variable nodes and returns a triple overlap score.

\paragraph{\wlk and \wwlk} \citet{opitz2021weisfeiler} apply the structural Weisfeiler-Leman kernel \cite{weisfeiler1968reduction, shervashidze2011weisfeiler} aiming at more contextualized AMR graph matches. The method extracts sub-graph statistics from the input graphs that describe different levels of node contextualizations. To assess a modulated similarity of AMR graphs, \citet{opitz2021weisfeiler} adapt the Wasserstein Weisfeiler-Leman metric \cite{NEURIPS2019_73fed7fd}, which compares the graphs in a joint latent space using the (permutation-invariant) Wasserstein distance.

\subsection{Aspectual AMR similarity}
\label{sec:focusedamrsim}

\paragraph{\fgsmatch: Fine-grained \smatch} \citet{damonte-etal-2017-incremental} create fine-grained \smatch-based metrics to analyze AMR similarity w.r.t.\ interesting semantic categories. We use \textbf{Frames}: graph similarity with regard to PropBank predicates. \textbf{Named entity}: graph similarity based on named entity substructures (\textit{person, city, ...}). \textbf{Negation}: graph similarity based on expressions of negation. \textbf{Concepts}: graph similarity based on node labels only. \textbf{Coreference}: graph similarity focused on co-referent structures. \textbf{SRL}: graph similarity considering predicate substructures. Finally, \textbf{Unlabeled}: not considering semantic edge labels.\footnote{We follow \citet{opitz-2020-amr} and set  metric values to 1.00 (as opposed to 0.00) in cases where neither of the graphs contains structures of the given aspect (e.g., named entities are absent from both graphs), since the graphs can then be considered to (vacuously) agree in the given aspect.} 

Additionally, we observe that AMR contains information about quantifiers and define \textbf{quantSim}, which measures the (normalized) overlap of quantifier structure of two AMRs. Although AMR lacks modeling of quantifier scope \cite{bos2016expressive}, estimating the overlap of quantificational structure can give indications of semantic sentence similarity.

\paragraph{Graph statistics} In addition, we introduce graph metrics that target other aspects modeled by AMR: \textbf{MaxIndegreeSim, maxOutDegreeSim} and \textbf{maxDegreeSim}. From each graph in a pair of AMRs, we extract the node that is best connected (either outdegree, indegree, or indegree+outdegree). We compare these nodes with cosine similarity using GloVe embeddings \cite{pennington-etal-2014-glove}. The motivation for this is that two Meaning Representations that share the same focus are more likely to be similar \cite{lambrecht1996information}. Similarly, \textbf{rootSim} compares the similarity of AMR roots, motivated by \citet{cai-lam-2019-core}, who speculate that more important concepts are closer to the root. 

\subsection{Data setup}

 For the decomposition objective  we need training instances of paired sentences with AMR metric scores attached. We proceed as follows:

1) We collect 1,500,000 sentence pairs from data sets that contain similar sentences.\footnote{AllNLI, CoCo, flickr captions, quora duplicate questions.} 2) We parse these sentences with a good off-the-shelf AMR parser.\footnote{\url{https://github.com/bjascob/amrlib} The parser is based on a fine-tuned T5 \cite{T5} language model and reports more than 80 Smatch points on AMR3. On a GPU Ti 1080 the parsing took approx.\ 3 weeks.} 3) For each training sentence pair we create a positive $(a, a^+)$ and a negative $(a, a^-)$ datum, where the negative pair is formed by replacing AMR $a^+$ with an AMR sampled from a random pair. Thereby we show \sbertx both AMR metric outputs computed from similar AMRs, and unrelated AMRs (that may still share some abstract semantic features). 4) We execute our AMR metrics (c.f.\ \S \ref{sec:genamrsim} \& \S\ref{sec:focusedamrsim}) over all pairs from step 3). Step 4) took approx.\ 3 days, since AMR metrics tend to have high computational complexity.

For experimentation, we cut off a development and testing set with 2,500 positive pairs each.\footnote{Using only similar sentence pairs for validation increases the AMR metric prediction difficulty and provides a useful lower bound for correlation.}

\section{Evaluation Study}

Our two objectives aim at creating \sbertx embeddings by partitioning \sbert's output space into features that capture different semantic AMR aspects, while controlling the decomposition process such that we prevent any forgetting of knowledge and preserve the power of the neural embeddings.

Hence, two key questions need to be addressed: 

\begin{enumerate}[label=\textbf{\arabic*.)}, noitemsep]
    \item Will \sbertx partition its sentence embedding space into interpretable semantic aspects?
    \item If so, what is the price? Does our consistency objective succeed in controlling the  decomposition process such that it retains \sbert's extraneous knowledge of sentence semantics?
\end{enumerate} 

\paragraph{Basic setup}  We use a standard \sbert model\footnote{\label{footnote-label}Pre-trained \texttt{All-MiniLM-L12-v2} from the sentence transformers library.} with 11 layers and allow tuning of the last two layers. The sentence embedding dimension is $d=384$, the sub-embedding dimension is set to $h=16$ for all 15 aspects of AMR, which implies that the dimension of the residual is $384-(15\times 16) = 144$. More details on the model architecture and the training hyper-parameters can be found in Appendix \ref{app:params}. In all result tables, $\dagger$ indicates statistically significant improvement over the runner-up (Student t-test, $p<0.05$, five random runs)

\subsection{\sbertx space partitioning} 

Our goal is to make \sbert embeddings more interpretable, by partitioning the sentence embedding space into multiple semantically meaningful sub-embeddings. We now aim to answer research question \textbf{ 1)} whether these sub-embeddings relate to the AMR metric aspects they were trained to predict.

\paragraph{Data setup} We use the 2,500 testing sentence pairs we had split from our generated data. For each semantic aspect, we calculate cosine similarities of the corresponding sub-embeddings. We then calculate the Spearmanr correlation of these predictions vs.\ the ground truth AMR metric similarities.  

\paragraph{Baseline setup} We consider three baselines. Same as \sbertx, all baselines are based on standard \sbert model.\textsuperscript{\ref{footnote-label}}

\textit{SB-full (no partitioning)}: We use the complete embedding, which means that we predict the same value for all AMR aspects. This baseline is bound to provide strong correlations with most metrics\footnote{Since AMR metrics correlate with human sentence similarity \cite{opitz2021weisfeiler}, and so does \sbert.}, but obviously lacks the interpretability we are aiming for. We therefore instantiate two more baselines that can be directly compared, since they partition the space according to semantic aspects. 

\textit{SB-rand (partitioning)}: We assign 16 embedding dimensions randomly to every semantic aspect. 

\textit{SB-ILP (partitioning)}: We use an integer linear program to assign the semantic aspects to different \sbert dimensions. We create a bi-partite weighted graph with node sets $(V_{SB}, V_{SEM})$ with \sbert dimensions ($V_{SB}$), and the targeted semantic aspects ($V_{SEM}$). Then, we introduce weighted edges $(i, j) \in V_{SB} \times V_{SEM}$, where a weight $\omega(i, j)$ is the Spearmanr correlation  of \sbert values in dimension $i$ vs.\ the metric scores for aspect $j$ across all (development) data instances. We solve (5--7).
\vspace{-1mm}
\begin{align}
    \max\sum_{(i,j) \in V_{SB} \times V_{SEM}} \omega(i, j)\cdot x_{ij} \\
    s.t.\ \sum_j{x_{ij}} \leq 1 ~\forall i \in V_{SB}\\
    \sum_i{x_{ij}} \geq 1 ~\forall j \in V_{SEM} 
\end{align}

The binary decision variables $x_{ij} \in \{0, 1\}$ indicate whether an \sbert dimension is part of a specific sub-embedding. The first constraint decomposes \sbert embeddings into non-overlapping parts, one for each aspect. The second constraint ensures that each semantic aspect is modeled.

\begin{table}
    \centering
    \scalebox{0.7}{
    \begin{tabular}{l|r|rrr}
    \toprule
        && \multicolumn{3}{c}{partitioning models} \\
        aspect & SB-full & SB-rand & SB-ILP & \sbertx \\ 
        \midrule
        \smatch & 64.6 &  57.1 & 57.9 & \textit{\textbf{68.2}}$^{\dagger}$  \\
        \wlk & \textit{76.7}$^{\dagger}$&  63.5 & 64.2 & \textbf{74.6} \\
        \wwlk & \textit{75.1}&  62.0 & 63.8 & \textbf{74.4} \\
        \midrule
        Frames & 46.0&  40.8 & 45.2 & \underline{\textit{\textbf{66.4}}}$^{\dagger}$ \\
        Unlabeled & 58.4 &  52.3 & 54.7 &\textit{\textbf{65.1}}$^{\dagger}$\\
        Named Ent.\ & -14.4 &  -1.1 & -0.3 & \underline{\textit{\textbf{51.1}}$^{\dagger}$}\\
        Negation & -2.00&  -0.0 & 3.4 & \underline{\textit{\textbf{33.0}}$^{\dagger}$}\\
        Concepts &\textit{ 76.7}$^{\dagger}$ &  64.5 & 72.3 & \textbf{74.0}\\
        Coreference & 23.2 &  10.3 & 13.6& \underline{\textbf{\textit{43.3}}}$^{\dagger}$\\
        SRL & 48.3 &  40.8 & 44.9 & \textit{\textbf{60.8}}$^{\dagger}$ \\
        maxIndegreeSim & 27.0 &  23.6 & 24.0 & \textit{\textbf{32.5}}$^{\dagger}$  \\
        maxOutDegreeSim & 22.3 & 17.5 & 19.4& \textit{\textbf{42.5}}$^{\dagger}$ \\
        maxDegreeSim & 22.3 &  18.0 & 19.7 & \textit{\textbf{30.0}}$^{\dagger}$ \\
        rootSim & 25.5 &  21.7 & 25.1 & \textit{\textbf{43.1}}$^{\dagger}$ \\
        quantSim & 11.5&  10.0 & 11.8 & \underline{\textit{\textbf{74.6}}$^{\dagger}$} \\
        \bottomrule
    \end{tabular}}
    \caption{Spearmanr x 100 of AMR aspects. \textit{Italics}: overall best. \textbf{bold}: best partitioning approach. \underline{underlined}: improvement by more than 20 Spearmanr points.} 
    \label{tab:predmetricresults}
\end{table}

\paragraph{Results} are displayed in Table \ref{tab:predmetricresults}. First, we see that the global AMR metrics \wlk and \wwlk are best modeled with the cosine distance computed on full \sbert embeddings (unpartitioned, Table \ref{tab:predmetricresults}) and we can't model them as well with a sub-embedding. This seems intuitive: the power of a low-dimensional sub-embedding is too low to express the complexity of the two Weisfeiler graph metrics that aim at capturing broader AMR sub-structures. However, the structural \smatch, which does not match structures beyond triples, can be better modeled in a sub-embedding (+3.8 vs.\ SB-full). Nonetheless, compared to the best partitioning baseline (SB-ILP), our approach provides substantial improvements (Spearmanr points, \wlk +10.4, \wwlk +10.6). 

Therefore, it is more interesting to study the  fine-grained semantic aspects measured by our aspectual AMR metrics. We find that there are three AMR features that are very poorly modeled with global \sbert embeddings: \textit{named entities}, \textit{negation}, \textit{quantification}. They also cannot be extracted with the SB-ILP baseline. By contrast, \sbertx clearly improves over these baselines. E.g., \textit{negation} modeling improves from a negative correlation to a significant positive correlation of 33.0 Spearmanr. \textit{Quantifier similarity} increases from 11.8 Spearmanr to 74.6.

\subsection{Correlation with human judgements}

Relating to research question \textbf{2)} on whether we can effectively prevent SBERT from forgetting prior knowledge when teaching it to predict AMR metrics, we test how well our approach compares to human ratings of sentence similarity in the typical zero shot setting. As  our main goal is to increase the interpretability of SBERT predictions, we consider \sbertx achieving SBERT's original performance on this task a satisfying objective.

\subsubsection{Sentence semantic similarity}

\paragraph{Test data} We use sentence semantic similarity data with human ratings. The STS (STSb) benchmark \cite{baudis-etal-2016-joint} assesses semantic similarity and SICK \cite{marelli-etal-2014-sick}  relatedness.\footnote{We min-max normalize the Likert-scale ratings of both datasets to the range between 0 and 1.}

\paragraph{Evaluation metric} We again use Spearmanr. To assess \textit{efficiency}, we display the approximate time for a metric to process 1,000 pairs. We also want to assess the \textit{explainability} of the methods, which can be complicated \cite{danilevsky-etal-2020-survey}. To keep it as simple as possible, we assign \xstar\xstar~when a metric is fully transparent and the score can be traced in the meaning space via graph alignment (\smatch, \wwlk), and \xstar~if there is a dedicated mechanism of explanation (e.g., via a linguistically decomposable score, as in \sbertx).

\paragraph{Baselines} As baselines we use: 1.\ \sbert and 2.\ our \sbertx from which we ablate a) the decomposition objective (\sbertx$^\text{\cancel{dec}}$) or b) the consistency objective (\sbertx$^\text{\cancel{cons.}}$). 
Assessing \sbertx$^\text{\cancel{cons.}}$ is key, since it shows the performance when we only focus on learning AMR features -- a significantly reduced score would prove the importance of counter-balancing decomposition with our consistency objective. For reference, we also include results from a simplistic baseline (word overlap) and the AMR metrics computed from the AMR graphs of sentences as in \citet{opitz2021weisfeiler}. 

\paragraph{Results} are shown in Table \ref{tab:stsresult}. Interestingly, while one main goal was to prevent a performance drop, \sbertx tends to outperform all baselines, including \sbert (significant improvement for STSb). 

It is important to note that catastrophic forgetting indeed occurs if learning is not controlled by the consistency objective. In this case, the performance drops by about 20-30 points (\sbertx$^\text{\cancel{cons.}}$ in Table \ref{tab:stsresult}). We conclude that our consistency objective effectively prevented any loss of embedding power.

\begin{table}
    \centering
    \scalebox{0.78}{
    \begin{tabular}{l|lrrr}
    \toprule
         system & speed (1k pairs) & xplain & STSb & SICK \\ 
         \midrule
        bag-of-words & 0s & - & 43.2 & 53.3\\
        bag-of-nodes & 31m (p) + 0.0s (i) & - & 60.4 & 61.6\\
        \smatch & 31m (p) + 49s (i) & \xstar\xstar & 57.2 & 59.1 \\
        \wlk & 31m (p) + 1s (i) & - & 63.9 & 61.4\\
        \wwlk & 31m (p) + 5s (i) & \xstar\xstar & 62.5 & 64.7\\
        \sbert & 1s (i) & - &  83.1 & 78.9\\
        \midrule
        \sbertx & 1s (i)  & \xstar & \textbf{83.7}$^\dagger$ & \textbf{79.1}\\
        \midrule
        \sbertx$^\text{\cancel{dec}}$ & 1s (i)   & - & 83.0 & 78.9\\
        \sbertx$^\text{\cancel{cons.}}$ & 1s (i)   & \xstar & 51.7 & 58.1\\
        \bottomrule
    \end{tabular}}
    \caption{Results on STSb and SICK using Spearmanr x 100; Speed measurements of parser (p) and metric inference (i), units are minutes (m) and seconds (s).}
    \label{tab:stsresult}
\end{table}

\begin{table}
    \centering
    \scalebox{0.79}{
    \begin{tabular}{l|r|r|lll}
    \toprule
    & &3-Likert& \multicolumn{3}{|c}{binary classif.\ F1 scores} \\
         system  &  xplain & Spea's r & Macro & Sim & $\neg$ Sim. \\ 
         \midrule
         RE19  &- &- & 65.4 & 52.3 & 78.5 \\
         BH21 & - & 34.8 & -& -& - \\
         OP21 & \xstar\xstar &- &  68.6 & 60.4 & 77.0 \\
        \sbert & - & 54.2 &71.7 & 63.8 & 79.6\\
        \midrule
        \sbertx & \xstar & \textbf{56.4$^\dagger$} &
        \textbf{72.9$^\dagger$} & \textbf{65.7$^\dagger$} & \textbf{80.1$^\dagger$} \\
        \midrule
        \sbertx$^\text{\cancel{cons.}}$ & \xstar & 28.2 & 55.6 & 53.7 & 57.4 \\
        \bottomrule
    \end{tabular}}
    \caption{Results on argument similarity prediction.}
    \label{tab:argresult}
\end{table}

\subsubsection{Argument similarity} 

\paragraph{Testing data} Besides the STS and SICK benchmarks we use the challenging UKPA(spect) data \cite{reimers-etal-2019-classification} with high-quality similarity ratings of natural language arguments from 28 controversial topics such as, e.g., \textit{GMO} or \textit{Fracking}.

\paragraph{Evaluation metric} Argument pairs in UKPA have one of four labels: \textit{dissimilar, unrelated, somewhat similar} and \textit{highly similar}. Originally, the task was evaluated as a binary classification task \cite{reimers-etal-2019-classification}, by mapping the \textit{similar} and \textit{highly similar} labels to 1, and the other two labels to zero. A similarity metric's scores are then mapped to binary decisions via a simple threshold-search script. To conform with this work, we also evaluate using this setup. But to account for the fine-grained labels, we also use a second metric based on (Spearmanr) correlation, following \citet{behrendt-harmeling-2021-arguebert} who propose a 3-Likert scale that maps \textit{dissimilar} and \textit{unrelated} to 0, \textit{somewhat similar} to 0.5, \textit{highly similar} to 1.0. 

\paragraph{Baselines} Table \ref{tab:argresult} shows the results of the best systems reported for i) a BERT-based approach  \cite{reimers-etal-2019-classification} (RE19), ii) the AMR-based \smatch-variant approach of \citet{opitz-etal-2021-explainable}, and iii) \citet{behrendt-harmeling-2021-arguebert} (BH21), who pre-train BERT on other argumentation datasets for 3-Likert style rating.

\paragraph{Results} \sbertx significantly outperforms all baselines, including \sbert, in the classification setting, and in the correlation evaluation setting. When assessing interpretability, OP21 offers \xstar\xstar~ because it is based on \smatch and the score can be \textit{fully} traced. However, it is less efficient, due to the cost of executing AMR metrics and parser, and lags behind in accuracy. Again, we can conclude that our approach offers a valuable balance between interpretability and performance. Finally, this experiment further corroborates that controlling the decomposition learning process is paramount: without consistency objective, the accuracy is almost halved (\sbertx$^\text{\cancel{cons.}}$ in Table \ref{tab:argresult}). 

\subsection{Ablation and parametrization experiments}

\paragraph{Upper-bounds for AMR metric approximation} While not the main objective of our work, the approximation of computationally expensive AMR metrics can be considered an interesting task on its own. We hence explore two AMR metric approximation upper-bounds: i) \textit{\sbertx$^\text{\cancel{cons.}}$}: Naturally, the consistency objective is orthogonal to the AMR metric approximation objective and by ablating the consistency objective, we can obtain an upper-bound for the prediction of AMR metric scores. \textit{ii) \sbertx$^\text{\cancel{cons.}}$+parser}: At the cost of making our approach much less efficient, we train \sbertx$^\text{\cancel{cons.}}$ directly on (linearized) AMR graph strings instead of their underlying sentences, which allows us to infer metric scores directly from AMR graphs. 

The results of these setups are given in Table \ref{tab:predmetricresults_upperbounds} in Appendix \ref{app:predmetricresults_upperbounds}. We see that both modifications can yield, to some extent, better AMR metric approximation accuracy, across all tested aspects. However, considering our second key goal of preserving the overall power of sentence embeddings, it is important to note that these improvements come at great cost, because if we do not control the decomposition process with our consistency objective, the similarity rating effectivity of the neural embeddings deteriorates (see \sbertx$^\text{\cancel{cons.}}$ in Table \ref{tab:stsresult} for sentence similarity and Table \ref{tab:argresult} for argument similarity). On top of this, \sbertx$^\text{\cancel{cons.}}$+parser will also lose much \textit{efficiency}.\footnote{Due to slow AMR parsing (c.f.\ Table \ref{tab:stsresult}).} 

\paragraph{Effect of parser quality} For creating AMRs, we used a strong parser that yields high \smatch scores on AMR benchmarks. To investigate the effect of using another parser, we re-ran our first experiment (decomposition) with metrics computed from parses of the older JAMR \cite{flanigan-etal-2014-discriminative} parser, that achieves more than 20 points lower \smatch on AMR benchmarks. We observe moderately(+1-3 correlation points) better results across all categories with the more recent parser. This implies that there is potential room for further improvement of our method by using an even more accurate parser, but judging from the marginally lower score of JAMR, the gain may be small.

\paragraph{Size of training data} We observe that the AMR metric approximation accuracy profits from growing size of the training data (see Appendix \ref{app:datsizes}).

\section{Data analyses with \sbertx}

\subsection{Studying \sbertx predictions}

We find many interesting cases where \sbertx is able to explain its similarity scores.\footnote{See more examples in Table \ref{tab:exanalysis}, Appendix \ref{app:exanalysis}.} For example, both \sbertx and \sbert assign a high similarity score (0.70--0.73) to \textit{two cats are looking at a window} vs.\ \textit{a white cat looking out of a window}, while the human similarity rating is just above average (.52). Here, a low similarity rating of -0.15 in \sbertx's \textbf{quantifier feature} provides a (possible) rationale for the much lower human score, due to a strong contrast in quantifier meaning (\textit{two} vs. \textit{a}). 

When confronted with \textbf{negation}, both \sbert and \sbertx assign moderately high scores to \textit{The man likes cheese} vs.\ \textit{the man doesn't like cheese}. But \sbertx can explain this: its high \textit{concept} similarity score increases the overall rating, while a (very) low similarity score for \textit{negation} (-0.30) regulates the rating downwards. We also see differences in how negation of a matrix verb affects the \sbertx negation feature -- compared with negation applied to a sub-ordinate sentence. \textit{Three boys in karate costumes \underline{[aren’t | are]} fighting} results in lower negation agreement  (Negation feature similarity: -0.31) compared to negation applying to the predicate of a sub-ordinate sentence, as in 
\textit{A child is walking down the street and a jeep \underline{[is not | is]} pulling up} (Negation feature similarity: -0.22).

\textbf{Coreference} can also explain key differences in meaning: \textit{The cat scratches a cat} and \textit{The cat scratches itself} are highly rated in all aspects (0.78--0.8 overall similarity) -- except for coreference, with   similarity of only 0.41, signaling a key difference reflected in coreference structures.

\begin{table*}[t]
    \centering
    \scalebox{0.62}{\begin{tabular}{l|l||rrrrrrrrrrr|rrrr|r}
    \toprule
     & & \multicolumn{11}{c}{aspectual semantic feature} & \multicolumn{4}{|c|}{global AMR feature} \\
     \midrule
        FEASIM  & data & Conc.\ & Frame & NE & Neg.\ & Coref & SRL & IDgr & ODgr & Dgr & $\sqrt{Sim}$ & quant & Sma.\ & Unlab.\ & WLK & W$^2$LK & Resid.\ \\
        vs.\ HUM & STSb & \textbf{73.8}$_{(1)}$ & 68.7 & 60.4 & 53.6 & 65.6 & 70.8$_{(2)}$ & 66.8 & 64.8 & 69.9$_{(3)}$ & 67.2 & 51.6 & 72.7 & 68.1 & \textbf{75.1} & 72.8 & \textbf{83.3}\\
        vs.\ SIM & STSb & \textbf{88.3}$_{(1)}$ & 81.5 & 75.6 & 61.9 & 80.0 & 84.4$_{(2)}$ & 81.2$_{(3)}$ & 78.7 & 81.2$_{(3)}$ & 77.5 & 60.1 &   86.1 & 83.4 & \textbf{88.9} & 86.4 & \textbf{99.3}\\
        vs.\ HUM & UKP & 51.3 & \textbf{61.3}$_{(1)}$ & 26.9 & 52.1$_{(3)}$ & 42.9 & 43.7 & 33.6 & 57.1$_{(2)}$ & 42.0 & 45.4 & -4.2 & 30.3 & \textbf{37.8} & 10.9 & 25.2 & 26.1 \\
        vs.\ SIM & UKP & \textbf{98.3}$_{(1)}$ & 86.7 & 85.0 & 93.3$_{(2)}$ & 91.7 & 90.0  & 90.0 & 91.7$_{(3)}$ & 85.0 & 86.7 & 63.3 & \textbf{91.7} & 86.7 & 81.7 & 86.7 & \textbf{96.7} \\
        \bottomrule
    \end{tabular}}
    \caption{Similarity investigation with \sbertx feature analysis. \textbf{bold}/(n): best from a feature group (rank 1--3).}
    \label{tab:feaana}
\end{table*}

Comparing the \textbf{foci of sentences} can also provide explanatory information. E.g., the human score for \textit{a man is smoking} and \textit{a baby is sucking on a pacifier} is zero, indicating complete dissimilarity. But \sbertx and \sbert assign scores that indicate moderate similarity. \sbertx's features may explain this, in that the sentences' foci (root sim) are somewhat related (0.4, \textit{smoking} vs.\ \textit{sucking}).

\subsection{Studying predictors of human scores}

What features can predict \textit{human similarity scores} and how may the assessment of argument similarity as opposed to sentence similarity differ from each other? In search for answers to these questions, we perform a quantitative analysis of \sbertx's fine-grained features. We proceed as follows: Let \textit{SIM} be \sbertx's similarity ratings for a pairwise data set, and \textit{HUM} be the corresponding human ratings. Now, let \textit{FEASIM} be the fine-grained \sbertx feature similarities for a feature \textit{FEA} (e.g., SRL aspect). Then we compute, for each \textit{FEA}, \textit{Spearmanr(FEASIM, SIM)} and 
\textit{Spearmanr(FEASIM, HUM)}, both on STS and argumentation benchmarks. In other words, we analyze predictive capacity of features for a) system vs.\ b) human similarity in c) different domains/tasks.

Analysis results are shown in Table \ref{tab:feaana}. Interestingly, for \textit{human argument similarity}, the residual has much lower predictive power (26.1), suggesting that human argument similarity notions differ significantly from sentence similarity. Indeed, another key difference can be found in the importance of quantification similarity, which is marginal (-4.2) for argumentation, but not for STS (51.6). We speculate that users judging argument similarity tend to generalize over  quantifier differences, being more focused on general statements and concepts, as opposed to, e.g., numerical precision. Notably, human argument similarity is markedly well predicted by \textbf{Frames} -- this feature alone achieves state-of-the-art results, indicating a marked importance of predicate frames for argument similarity. 

Of course, although the analysis may give some interesting indications about similarity as perceived by humans (and \sbert), it has to be taken with a grain of salt, one reason being, e.g., that the shown statistics are influenced by AMR metric prediction accuracy, which varies across aspects (c.f.\ Table \ref{tab:predmetricresults}). Our study also indicates that neither sentence nor argument similarity can be fully explained by any feature. We hypothesize that we may need to go beyond what \sbert and (current) AMR metrics can measure, e.g., by incorporating background knowledge. Our method may offer a way to inject such background knowledge into sentence embeddings, via distillation of dedicated metrics. 

\section{Conclusion}

We propose a method for decomposing neural sentence embedding spaces into different sub-spaces, with the goal of obtaining sentence similarity ratings that are \textit{accurate, efficient} and \textit{explainable}.  The sub-spaces express facets of meaning as captured by AMR and AMR metrics, such as \textit{Negation} or \textit{Semantic Roles}. The \textit{decomposition objective} partitions the semantic space via targeted synthesis of AMR metrics. The effectiveness of neural sentence embeddings is preserved by a \textit{consistency objective} that controls the decomposition process and routes global semantic information not expressed by AMR into a \textit{residual embedding}. The S$^3$BERT embeddings are more explainable and are on par, or even outperform, SBERT's accuracy. Our approach allows straightforward extension to customized metrics of meaning similarity.

\section*{Acknowledgements}

We are grateful to our anonymous reviewers for their comments that have helped to significantly improve our paper. We also thank Julius Steen and Philipp Wiesenbach for helpful feedback on ideas.
The work has been partially funded by the DFG through the project ACCEPT as part of the Priority Program “Robust Argumentation Machines” (SPP 1999).

\bibliography{anthology,custom}
\bibliographystyle{acl_natbib}

\appendix

\section{Appendix}
\label{sec:appendix}

\subsection{Hyper-parameters and training}
\label{app:params}

Batch size is set to 64, the learning rate (after 100 warm-up steps) is set to 0.00001. We train for 8 epochs, evaluating every 1000 steps. Afterwards we select the model from the evaluation step where we achieve minimum development loss. 

\subsection{Scaling training data size}
\label{app:datsizes}

See Table \ref{tab:datsizes}.

\begin{table}
    \centering
    \scalebox{0.7}{
    \begin{tabular}{l|r|rrr}
    \toprule
    & \multicolumn{4}{c}{amount of training data}\\
        aspect & rand (0k) & 50k& 300k &1500k \\ 
        \midrule
        \smatch & 57.1& 59.4  & 60.2 &68.2  \\
        \wlk & 63.5& 64.1   & 70.2 &74.6 \\
        \wwlk & 62.0&  65.8  & 67.0 & 74.4 \\
        \midrule
        Frames &40.8 &44.2  & 53.6 & 66.4 \\
        Unlabeled & 52.3& 53.6  &54.1 & 65.1\\
        Named Ent.\ & -1.1&  11.4  & 31.8 & 51.1\\
        Negation &-0.0&  17.8  &29.0 & 33.0\\
        Concepts  & 76.7& 69.6  & 71.2& 74.0  \\
        Coreference & 23.2&  23.9 & 25.2 & 43.3\\
        SRL & 48.3& 49.4  & 50.0 & 60.8\\
        maxIndegreeSim &27.0 & 26.7 &26.4 & 32.5  \\
        maxOutDegreeSim & 22.3 & 22.4 &23.1  & 42.5 \\
        maxDegreeSim & 22.3& 22.1  &22.5 &30.0 \\
        rootSim & 25.5 & 26.4 &28.9  & 43.1 \\
        quantSim & 11.5 & 47.1 & 65.4 &74.6 \\
        \bottomrule
    \end{tabular}}
    \caption{AMR prediction performance w.r.t. different training data sizes.} 
    \label{tab:datsizes}
\end{table}

\subsection{AMR metric approximation upper-bounds}
\label{app:predmetricresults_upperbounds}

See Table \ref{tab:predmetricresults_upperbounds}.

\begin{table}
    \centering
    \scalebox{0.65}{
    \begin{tabular}{l|r|rr}
    \toprule
        & \\
        aspect & \sbertx & \sbertx$^\text{\cancel{cons.}}$ & \sbertx$^\text{\cancel{cons.}}$+parser \\ 
        \midrule
        \smatch & 68.2 & 77.0 & 80.3\\
        \wlk &74.6 & 79.3 & 78.9\\
        \wwlk & 74.4 & 81.5 & 82.3\\
        \midrule
        Frames & 66.4 & 79.6 & 80.3\\
        Unlabeled & 65.1 & 75.5 & 78.0 \\
        Named Ent.\ & 51.1 & 58.0 & 61.9\\
        Negation & 33.0 & 34.5 & 35.5\\
        Concepts & 74.0 & 78.5 & 76.4\\
        Coreference & 43.3 & 57.4 & 72.1\\
        SRL & 60.8 & 74.3 & 83.0\\
        maxIndegreeSim & 32.5 & 37.3 & 37.5 \\
        maxOutDegreeSim & 42.5 & 59.9 & 65.4\\
        maxDegreeSim & 30.0 &40.6 & 42.7 \\
        rootSim & 43.1 & 57.4 & 81.2\\
        quantSim & 74.6 & 75.7 &76.1 \\
        \bottomrule
    \end{tabular}}
    \caption{AMR metric approximation upper-bounds.  \textit{\sbertx$^\text{\cancel{cons.}}$}: \sbertx without consistency objective (trades sentence similarity rating performance for better AMR approximation). \textit{\sbertx$^\text{\cancel{cons.}}$+parser}: \sbertx without consistency objective and inference on linearized AMR graphs (trades sentence similarity rating performance \textit{and} efficiency for better AMR approximation). } 
    \label{tab:predmetricresults_upperbounds}
\end{table}

\subsection{Prediction examples}
\label{app:exanalysis}

See Table \ref{tab:exanalysis}.

\begin{table*}[ht!]
    \centering
    \scalebox{0.64}{
    \begin{tabular}{llllll}
    \toprule
        index &  sentence pairs  & humSim & \sbert & \sbertx & notable feature similarities \\
         \midrule
         	
\rowcolor{LightCyan}
      1 &  two cats are looking at a window &   0.52 & 0.70& 0.72 & concepts: 0.87$\uparrow\uparrow$; quant: -0.15$\downarrow\downarrow$ \\
      	
\rowcolor{LightCyan}
       & a white cat looking out of a window &&&& \\
       
      2 & three men posing in a tent  & 0.24 & 0.39&0.42 & quant:0.99$\uparrow\uparrow$; Frames: -0.02$\downarrow\downarrow$, Unlabeled: 0.6 $\uparrow$  \\
      & three men eating in a kitchen \\
      	
\rowcolor{LightCyan}
       3&  rocky and apollo creed are running down the beach  & 0.6 & 0.33& 0.32 & maxDegSim: 0.4$\uparrow$, NamedEnt: -0.72$\downarrow\downarrow$  \\
       	
\rowcolor{LightCyan}
       & the men are jogging on the beach&&&& \\
       
       4 & a man is smoking  & 0.0 & 0.06 & 0.06 & rootSim$\uparrow\uparrow$: 0.4 \\
       & a baby is sucking on a pacifier \\
       	
\rowcolor{LightCyan}
 5 & a dog prepares to herd three sheep with horns & 0.44 & 0.63& 0.65 & SRL: 0.56$\downarrow$; Frames: 0.45$\downarrow$, Concepts: 0.85$\uparrow$\\
 	
\rowcolor{LightCyan}
 & a dog and sheep run together&&&&  \\
 
6  & The cat scratches itself  & na& 0.81&0.78 & Concepts: 0.9 $\downarrow$; Negation 0.56$\downarrow$; Coref: 0.41$\downarrow\downarrow$ \\
&  The cat scratches another cat \\
	
\rowcolor{LightCyan}
7 & The man likes cheese & na & 0.80 & 0.77 & Concepts: 0.90 $\uparrow$; Negation: -0.3 $\downarrow\downarrow$ \\
	
\rowcolor{LightCyan}
 & The man doesn't like cheese &&&&\\

8 & Recruits are talking to an officer & 0.68 & 0.97 & 0.98 & SRL: 0.96 $\downarrow$; Negation: 0.90 $\downarrow$; Unlabeled: 0.99 $\uparrow$ \\
 & An officer is talking to the recruits \\
 	
\rowcolor{LightCyan}
9 & A dog is teasing a monkey at the zoo & 0.63 & 0.99 & 0.99 & SRL: 0.96 $\downarrow$; Negation: 0.97 $\downarrow$; maxDegr: 1.0 $\uparrow$\\
	
\rowcolor{LightCyan}
& A monkey is teasing a dog at the zoo &&&& \\

10 & Three boys in karate costumes aren't fighting & 0.58 & 0.86&0.86 & Concepts: 0.92$\uparrow$; Negation: -0.31$\downarrow\downarrow$ \\
& Three boys in karate costumes are fighting \\
\rowcolor{LightCyan}
11 & A child is walking down the street and a jeep is pulling up  & 0.63& 0.95 &  0.92 & Concepts: 0.95$\uparrow$; Negation: -0.22$\downarrow\downarrow$ \\
\rowcolor{LightCyan}
& A child is walking down the street and a jeep is not pulling up &&&& \\
 \bottomrule
    \end{tabular}}
    \caption{Prediction Examples from STSb and SICK, or own construction (human rating: na).}
    \label{tab:exanalysis}
\end{table*}

\end{document}